\documentclass[runningheads]{llncs}
\usepackage[T1]{fontenc}
\usepackage{graphicx}
\usepackage{float}
\usepackage{eccvabbrv}
\usepackage{amsmath}
\usepackage{amsfonts}
\usepackage{bbding}
\usepackage{booktabs}
\usepackage{multirow}
\usepackage{multicol}
\usepackage{hyperref}
\usepackage[capitalize]{cleveref}
\usepackage{color}
\usepackage[table]{xcolor}

\begin{document}
\title{Tuning Vision-Language Models with Candidate Labels by Prompt Alignment}
\titlerunning{Tuning VLMs with Candidate Labels by Prompt Alignment}


\author{
Zhifang Zhang\inst{1}\and
Yuwei Niu\inst{2}\and
Xin Liu\inst{3}\and
Beibei Li \inst{2}\thanks{Corresponding author}
}

%
\institute{University of Queensland \and
Chongqing University \and
Southeast University
}
%
%
\maketitle              
%
\begin{abstract}
Vision-language models (VLMs) can learn high-quality representations from a large-scale training dataset of image-text pairs. 
Prompt learning is a popular approach to fine-tuning VLM to adapt them to downstream tasks.
Despite the satisfying performance, a major limitation of prompt learning is the demand for labeled data. In real-world scenarios, we may only obtain candidate labels (where the true label is included) instead of the true labels due to data privacy or sensitivity issues.
In this paper, we provide the first study on prompt learning with candidate labels for VLMs. We empirically demonstrate that VLMs can learn from candidate labels through prompt learning.
Nonetheless, its performance drops when the label ambiguity increases. In order to improve its robustness, we propose a simple yet effective framework that better leverages the prior knowledge of VLMs to guide the learning process with candidate labels. 
Specifically, our framework disambiguates candidate labels by aligning the model output with the mixed class posterior jointly predicted by both the learnable and the handcrafted prompt. Besides, our framework can be equipped with various training objectives for learning with candidate labels to further improve their performance. Extensive experiments demonstrate the effectiveness of our proposed framework.


\end{abstract}
\section{Introduction}
\label{sec:intro}

Vision-language models (VLMs) such as CLIP \cite{radford2021clip}, ALIGN \cite{pmlr-v139-align}, and Coca \cite{yu2022coca} have become excellent base models in multiple domains, most of which employ a dual-encoder architecture to align the natural images with descriptive texts. 
Remarkably, this special training pattern has endowed VLMs with superior zero-shot transfer performance on visual recognition tasks. In specific, during the inference, the pre-trained text encoder receives inputs in the form of man-crafted prompts, \eg, \textit{``a photo of \textless{}CLS\textgreater{}.''}. Subsequently, all the generated textual embeddings are matched with the visual embedding obtained from the image encoder to predict the image category. 
However, the powerful zero-shot ability of VLMs was shown to be heavily dependent on the choice of handcrafted prompts, which needs substantial efforts and professional domain knowledge to design \cite{zhou2022coop}.
To avoid the manual design of the prompts, \textit{prompt learning} \cite{zhou2022coop} is proposed, which treats the textual prompt as additional learnable parameters and tunes them while keeping all the original parameters of the pre-trained model fixed. Later, the concept of the prompt is extended to visual prompt \cite{jia2022visual} and multi-modal prompt \cite{khattak2023maple} in VLMs. 
Overall, there has been increasing attention paid to prompt learning due to its potential to perform significantly better than zero-shot transfer with a few sets of labeled data \cite{zhou2022coop,zhou2022cocoop,zhu2023gradalign,khattak2023maple,jia2022visual}.
\begin{figure}[!t]
    \centering
    \includegraphics[width=0.8\linewidth]{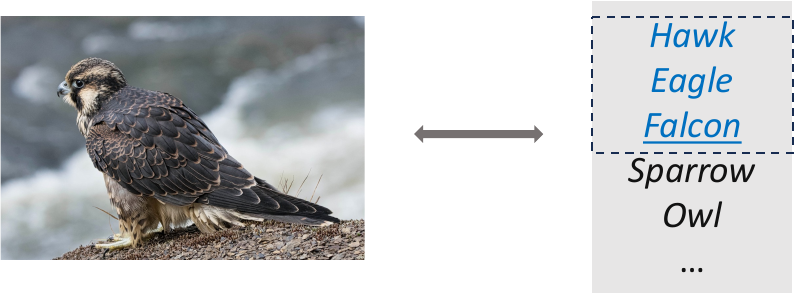}
    \caption{There are possible candidate labels corresponding to the image of a falcon. In the right portion of the figure, candidate labels are colored in blue. In the candidate label set, the hidden true label is underlined to be distinct from the false-positive labels.}
    \label{fig:cl_illustration}
\end{figure}

While prompt learning has demonstrated effectiveness and efficiency in few-shot supervised learning, the true labels must be provided for the training data used in prompt learning. 
This is a significant defect and will limit the usage of prompt learning in various real-world scenarios because we may be unable to collect accurate labels due to security issues or labeling difficulties. Fortunately, obtaining a set of candidate labels that includes the true label in these situations is easier.
For example, as shown in Figure \ref{fig:cl_illustration}, it is challenging to determine which is the true label from `Hawk', `Eagle', and `Falcon', hence all these three labels can be considered as candidate labels during the annotation process of this bird recognition task.
As we see, learning with only candidate labels (also widely known as \textit{partial-label learning (PLL) } \cite{wang2021pico,wu2022revisiting,lv2020progressive,feng2020provably}) is practically significant, which also has arisen in many vital applications such as web mining \cite{luo2010learning}, online annotation \cite{tang2017confidence} and ecoinformatics \cite{liu2012conditional}. Nevertheless, existing PLL methods primarily focus on training a model from scratch, and the effectiveness of PLL in the new training paradigm called prompt learning remains unconfirmed. To bridge this research gap, we, for the first time, explore the validity and potential approaches for prompt learning with candidate labels.

This paper empirically shows that prompt learning combined with the prevailing PLL training objectives in a vanilla way can learn from candidate labels. However, as experimentally suggested, if the candidate labels become more ambiguous, the model's performance will drop significantly. Fortunately, prompt learning is still more robust than linear probe \cite{radford2021clip}, another tuning method that trains a linear classifier on top of a frozen pre-trained model. We conjecture that the reason lies in the prior knowledge brought by the fixed class token, which can keep VLMs from over-fitting to the false-positive labels in the candidate label set and provide VLMs with preferred zero-shot ability, thus mitigating the error accumulation problem \cite{yao2021network} in PLL when the label ambiguity increases.

Therefore, to enhance the robustness of prompt learning with candidate labels, we propose a simple yet effective framework incorporating the handcrafted prompt to distill the model with more comprehensive prior knowledge.
Concretely, it dynamically mixes the class posteriors predicted by both the handcrafted and learnable prompt, followed by aligning the mixed class posterior with the model output.
Besides, due to the simplicity and flexibility of our framework, it can cooperate with any current PLL training objectives. With our framework, the overall performance of various PLL training objectives has improved by a large margin when tuning VLMs with candidate labels.

Our main contributions can be summarized as follows: 
\begin{itemize}
\item We provide the first study on the scenario when vision-language models are tuned with only candidate labels.
\item We demonstrate empirically and explain that prompt learning combined with PLL training objectives in a vanilla way can learn from candidate labels but is not robust when the label ambiguity is high.
\item A framework is proposed to enhance the robustness of prompt learning with candidate labels by aligning the model output with the dynamically mixed prediction by both the handcrafted and the learnable prompt.
\item Extensive experiments demonstrate the effectiveness of our framework.
\end{itemize}

\begin{figure*}[!tbp]
    \centering
    \includegraphics[width=1\linewidth]{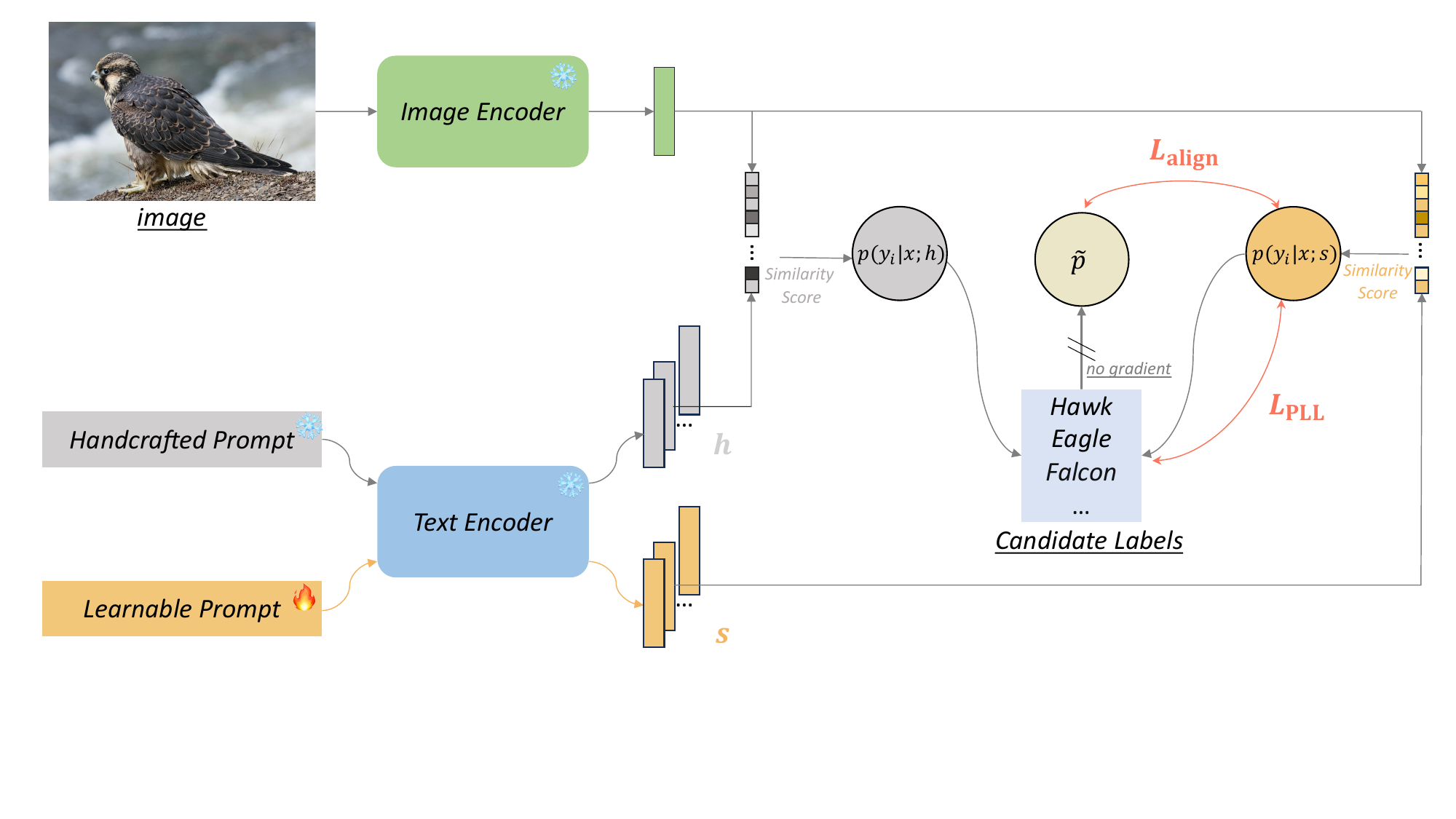}
    \vspace{-2cm}
    \caption{Illustration of our framework. Our framework includes a prompt alignment module and a PLL method module. Regarding the prompt alignment module, two different prompts are input to the text encoder, including the handcrafted prompt and the learnable prompt. Afterwards, we mix the recalculated class posteriors yielded by the handcrafted prompt and the learnable prompt. Then, the mixed class posterior is aligned with the output of the learnable prompt using the re-weighted cross-entropy loss. For the PLL method module, any prevailing PLL methods can be combined with the prompt alignment module. It is important to know that during fine-tuning, all the parameters of this framework are \textbf{frozen} except for the learnable prompt.}
    \label{fig:our_framework}
    \vspace{-0.6cm}
    
\end{figure*}

\section{Preliminaries of Prompt Learning and PLL}
\label{sec:preliminaries}

\subsection{Prompt Learning}
Recently, significant advancements have been made in the field of vision-language models (VLMs) \cite{pmlr-v139-align,radford2021clip,yu2022coca,zhai2022lit}. Unlike models learning from uni-modal supervision, VLMs align visual and textual signals to learn rich representation from massive image-text pairs in the pre-training stage. One of the keys to the excellent performance of VLMs is the tremendous amount of training data. CLIP \cite{radford2021clip} trains both the text and image encoder simultaneously using a contrastive loss on a dataset of 400 million image-text pairs. ALIGN \cite{pmlr-v139-align} leverages a noisy pre-training dataset of 1.8 billion pairs to train the model. 
However, although VLMs have shown promising performance in generalizing to new concepts without auxiliary information, their large amount of parameters makes it impractical to fine-tune the entire model \cite{lialin2023scaling}. Besides, full-parameter fine-tuning will also make large models prone to overfitting to downstream tasks and \textit{catastrophic forgetting} \cite{mccloskey1989catastrophic}. 
To address the above issues, multiple transfer learning methods \cite{zhou2022coop,gao2023clipadapter,radford2021clip,zhang2021tip} are proposed to adapt VLMs to downstream tasks effectively and efficiently. Linear probe \cite{radford2021clip} freezes the pre-trained VLM and trains a linear classifier on top of the image encoder. CLIP-adapter \cite{gao2023clipadapter} introduces an adapter with residual-style feature blending with the pre-trained features. Notably, most fine-tuning methods heavily depend on the acquisition of precisely labeled data.

Among all fine-tuning methods that adapt VLMs to a new task, one typical approach is prompt learning. It treats the text prompt, \eg, \textit{``a photo of a \textless{}CLS\textgreater{}''} as continuous learnable parameters, and optimizes the prompt with multiple vision tasks, including image classification 
\cite{zhou2022coop,khattak2023maple}, dense prediction \cite{rao2022denseclip}, `. CoOp \cite{zhou2022coop} is the first work that migrated prompt learning to vision tasks, which tunes CLIP by optimizing the parameters of the learnable textual prompt (also called soft prompt) while keeping the class token fixed.

To be specific, assume CoOp introduces $M$ learnable vectors $\{\boldsymbol{v}_k \}_{k=1}^M$ and $C$ fixed class tokens $\{ \boldsymbol{c}_l\}_{l=1}^C$. Together, they are usually concatenated to form the full prompt $\boldsymbol{s}_i = \{\boldsymbol{v}_1, \boldsymbol{v}_2, ..., \boldsymbol{v}_M, \boldsymbol{c}_i\}$ for class $i$. Let the normalized image embedding be $\boldsymbol{f}^v$, then the class posterior is estimated as:
\begin{equation}
    p(y=i|\boldsymbol{x}) = \frac{\mathrm{exp}(\mathrm{sim}(\boldsymbol{f}^v,\mathrm{TextProj}(\boldsymbol{s}_i))/\tau )}
{ {\textstyle \sum_{j=1}^{C}} \mathrm{exp}(\mathrm{sim}(\boldsymbol{f}^v, \mathrm{TextProj}(\boldsymbol{s}_j))/\tau)}. 
\label{coop fewshot}
\end{equation}
At last, the learnable context vectors $\{\boldsymbol{v}_k \}_{k=1}^M$ are optimized on a dataset $D=\{ (\boldsymbol{x}_i,y_i)_{i=1}^N \}$ with the cross-entropy loss: 

\begin{equation}
    \mathcal{L}_{\mathrm{true}} = - \mathbb{E}_{(\boldsymbol{x},y) \in D}[\log p(y|\boldsymbol{x})].
    \label{coop fewshot loss}
\end{equation}

Notably, optimizing this training objective of CoOp requires examples of true labels. But in many realistic scenarios, the true label is not accessible due to multiple reasons. 
The former research has focused on prompt learning with noisy labels \cite{noiseprompt} or no label \cite{huang2022upl,zhang2024candidate} and showed prompt learning is not only robust with label noise but also can effectively learn from unlabelled data with specifically designed algorithms.

In this work, we focus on the scenario where only a candidate label set can be obtained, which we unfortunately cannot utilize to optimize the above training objective.
Therefore, we will study prompt learning with candidate labels using PLL training objectives instead of cross-entropy loss.

\subsection{Partial-label Learning}
\label{sec:preliminaries:pll}
Partial-label learning (PLL) allows each training example to be annotated with a candidate set, containing the true label. 
The problem is defined as follows: In the training process, we can obtain a candidate label set $Y_i $ for sample $x_i$ that contains the true label $y_i$, \ie, $y_i \in Y_i$. 
To simplify the problem, in the following experiments, unless otherwise stated, we assumed that the the candidate labels in partial label datasets manually constructed (also called `partialized') from fully-supervised datasets for PLL method evaluation are generated uniformly by flipping negative labels $\Bar{y}_i \ne y_i$ to false-positive labels inside the candidate label set with label ambiguity $q = P(\bar{y}_i \in Y_i |\bar{y}_i \ne y_i )$. 
Two mainstream strategies have been developed to address this problem: the \textit{averaged-based} strategy and the \textit{identification-based} strategy. Average-based strategy disambiguates the candidate labels by treating them equally \cite{hullermeier2006learning,cour2011learning,zhang2015solving}. 
However, the main drawback of this strategy is that the learner can be seriously affected by multiple false-positive labels \cite{lv2023robustness}.
Identification-based strategy regards the true label as a latent variable and selects the most likely label for training \cite{jin2002learning,feng2018leveraging,nguyen2008classification,feng2019partial}, which is prone to error accumulation if the wrong label is selected initially \cite{yao2021network}. 
As deep learning thrives, many PLL algorithms have been proposed for training with deep neural network \cite{feng2020provably,lv2020progressive,lv2023robustness,zhang2021exploiting}. Feng \etal \cite{feng2020provably} assumes the generation process of partial labels and derives classifier-consistent and risk-consistent methods from a theoretical perspective. PRODEN \cite{lv2020progressive} updates the model parameters and identifies the true labels seamlessly. PiCO \cite{wang2021pico} divides the learning process into representation learning by contrastive loss and label disambiguation by prototype and pseudo-target updating. 
In particular, as the scale of the parameters and training data of modern deep neural expands \cite{devlin2018bert,brown2020gpt3,kirillov2023segment}, the deep learning community has embraced a new training paradigm of \textit{pre-training and fine-tuning}. 
Recently, Zhang \etal \cite{zhang2024candidate} proposed a strategy that allows VLMs to generate candidate pseudolabels to enhance prompt learning with unlabelled data. 
Their work is orthogonal to ours, as we enhance prompt learning performance with human-annotated candidate labels, while they focus on using unlabelled data.

\vspace{-0.3cm}
\section{Prompt Learning with Candidate Labels}
\vspace{-0.3cm}

Prompt learning has been shown to be effective when tuning VLMs, provided a few examples with perfect labels. Nevertheless, it remains unclear how prompt learning performs with candidate labels. 

In this section, we empirically demonstrate the viability of it combined with PLL training objectives in a vanilla way for learning with candidate labels and explain the reasons.
Then, based on the observations above, we propose a framework that dynamically aligns the mixed class posterior with the model's output to enhance performance.
Figure \ref{fig:our_framework} depicts our framework.

\subsection{Pilot Experiments}
\label{pilot}

\begin{figure*}[!htbp]
    \centering
    \includegraphics[width=1\linewidth]{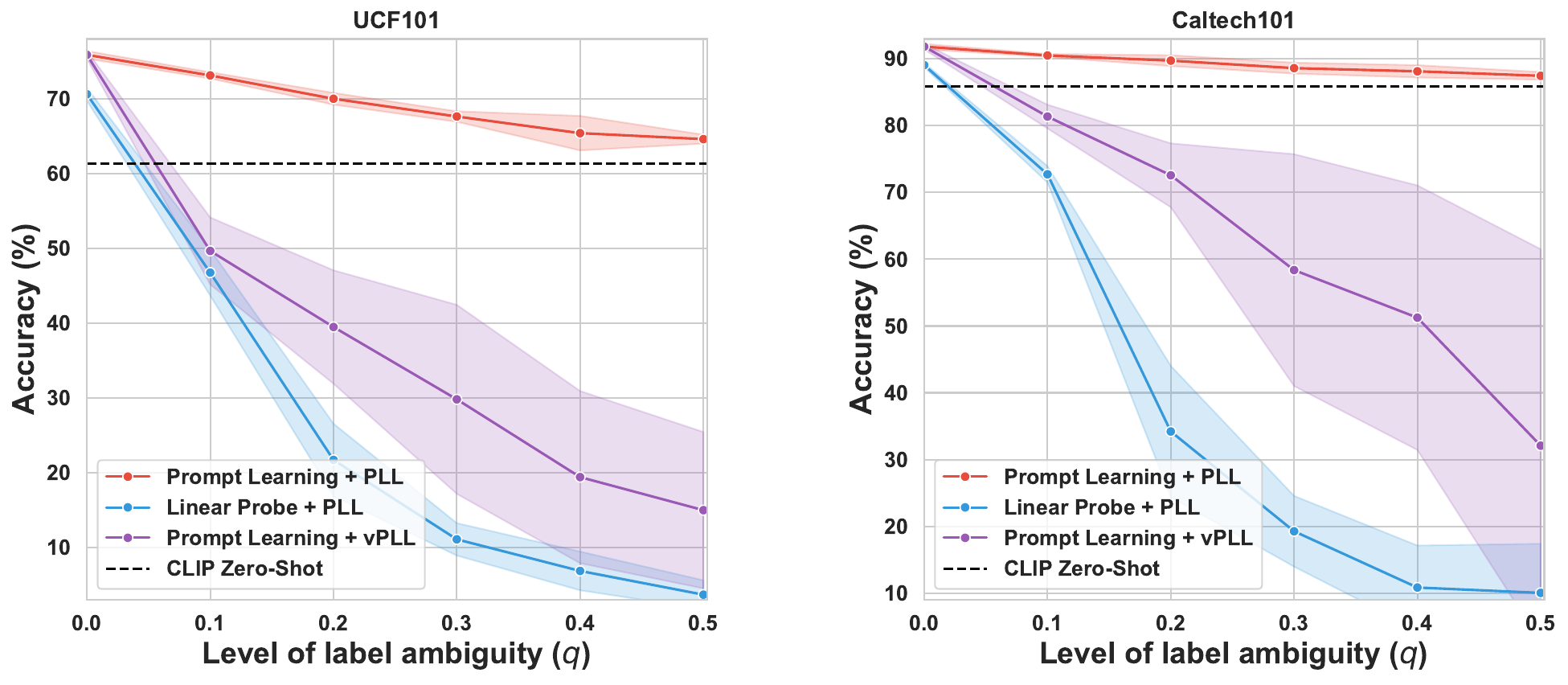}
    \caption{Performance comparison with multiple fine-tuning approaches combined with PiCO \cite{wang2021pico} in a vanilla way on UCF101 \cite{soomro2012ucf101} and Caltech101 \cite{fei2004learning} with candidate labels of the incremental label ambiguity. vPLL means a simple baseline that treats every candidate label as the ground-truth label and uses cross-entropy loss to learn.}
    \label{fig:C1}
\vspace{-0.6cm}
\end{figure*}

This part explores prompt learning with PLL training objectives and presents our key findings.
Linear probe \cite{tian2020probe,radford2021clip} is a strong baseline in few-shot learning, which trains a linear classifier on top of the pre-trained frozen image encoder of VLMs. 
PiCO \cite{wang2021pico}, a prevailing PLL training objective, uses embedding prototypes to identify the ground-truth label from the candidate set and train the classifier, which in turn selects a positive set to calculate the contrastive loss and better align the embedding prototypes.

In the pilot experiment, we use linear probe and prompt learning to tune the pre-trained model simply combined with PiCO on the uniformly partialized datasets of UCF101 \cite{soomro2012ucf101}, and Caltech101 \cite{fei2004learning}.

Moreover, we utilize ResNet-50 \cite{he2016resnet} as the same visual backbone of the two fine-tuning methods and compare their performance on the datasets of UCF101 and Caltech101 under different levels of label ambiguity. 
For linear probe, the learning rate is selected from \{2.5,1,0.5,0.25,0.1,0.05, 0.01\} to ensure optimal performance.
The other training settings for linear probe and prompt learning are consistent with the implementation details in the experiment section. 
Additionally, in the remaining content, `prompt learning combined with PLL training objectives in a vanilla way' is abbreviated as `vanilla prompt learning'.

The result is shown in Figure \ref{fig:C1}. We can see that vanilla prompt learning outperforms CLIP zero-shot, demonstrating the viability of learning with candidate labels through prompt learning. However, as the level of label ambiguity increases from $0$ to $0.5$, the performance deteriorates significantly (\textit{-4.4\%} for Caltech101 and \textit{-11.3\%} for UCF101), which is unacceptable.

On the other hand, it is shown that vanilla prompt learning is more robust with candidate labels than linear probe. We conjecture that it is the fixed class token that introduces the prior knowledge, enhancing the model's robustness. Specifically, the prior knowledge not only acts as a regularizer, keeping the model from over-fitting to the false-positive labels \cite{noiseprompt}, but also equips the model with a preferred zero-shot ability to identify the concealing true label better. These effects can jointly mitigate the error accumulation problem \cite{yao2021network} in PLL. 

Therefore, to enhance the model's robustness with candidate labels, inspired by the advantages of the fixed class token, we propose a framework that introduces the handcrafted prompt to distill the model with full prior knowledge.

\vspace{-0.2cm}
\subsection{Our Framework}

This framework is proposed to improve the robustness of prompt learning with candidate labels. It provides a powerful regularization that aligns the dynamically mixed prediction of the handcrafted and learnable prompt with the current model output using weighted cross-entropy loss.
It is shown in Figure \ref{fig:our_framework}. 

\noindent\textbf{Prompt Alignment Regularization.}\quad

Let $\mathcal{X}$ be the input space, and $\mathcal{Y} = \{1,2,\ldots,C\}$ be the label space. 
The $i$-th learnable prompt $\boldsymbol{s}_i = \{\boldsymbol{v}_1, \boldsymbol{v}_2, ..., \boldsymbol{v}_M, \boldsymbol{c}_i\}$, where
$\{\boldsymbol{v}_m \}_{m=1}^M$ denotes $M$ learnable tokens and $\boldsymbol{c}_i$ is the the word embedding for the $i$-th class name.
Similarly, the $i$-th manually crafted prompt is denoted as $\boldsymbol{h}_i$. 
To clarify, $f_i(\boldsymbol{x};\boldsymbol{s})$ and $g_i(\boldsymbol{x};\boldsymbol{h})$ are the softmax outputs of the $i$-th label, as predicted by the learnable prompt and handcrafted prompt separately. 
When $(\boldsymbol{x}, Y)$ is drawn from the partialized dataset, the prompt alignment loss can be calculated as: 
    \begin{equation}
\mathcal{L}_{\mathrm{align}}(\boldsymbol{x},Y) = - \sum\nolimits_{i=1}^{C}  \ \tilde{p}_{i}\log f_i(\boldsymbol{x};\boldsymbol{s}),
    \label{align loss}
\end{equation}
where $\tilde{p}_{i}$ is mixed linearly with the class posteriors predicted by both the learnable and handcrafted prompts: 
\begin{equation}
    \tilde{p}_{i} = \alpha p(y=i\mid \boldsymbol{x};\boldsymbol{s}) + (1-\alpha)p(y=i\mid \boldsymbol{x};\boldsymbol{h}).
    \label{mix}
\end{equation}
Since the non-candidate labels can never be the ground-truth label, the class posteriors are recalculated as: 
\begin{equation}
     p(y=i\mid \boldsymbol{x};\boldsymbol{s}) = 
\begin{cases}
\frac{f_i(\boldsymbol{x};\boldsymbol{s})}{\sum_{j \in Y}f_j(\boldsymbol{x};\boldsymbol{s})},
& \text{$i \in Y$},\\
0,& \text{$i \notin Y$}.
\end{cases}
\end{equation}
\begin{equation}
     p(y=i\mid \boldsymbol{x};\boldsymbol{h}) = 
\begin{cases}
\frac{g_i(\boldsymbol{x};\boldsymbol{h})}{\sum_{j \in Y}g_j(\boldsymbol{x};\boldsymbol{h})} ,
& \text{$i \in Y$},\\
0,& \text{$i \notin Y$}.
\end{cases}
\end{equation}
This regularization can be adapted to any prevailing PLL training objective:
\begin{equation}
\mathcal{L}_{\mathrm{total}}(\boldsymbol{x},Y) =  \mathcal{L}_{\mathrm{PLL}}(\boldsymbol{x},Y) + \beta\mathcal{L}_{\mathrm{align}}(\boldsymbol{x},Y),    
\end{equation}
where $\beta$ is the factor controlling the strength of the alignment loss.

\noindent\textbf{Dynamic Mixing Strategy.}\quad

In Equation (\ref{mix}), we use a balancing factor $\alpha$ to mix the handcrafted and learnable prompt predictions. 
However, a fixed balancing factor may be sub-optimal since the prediction quality of the handcrafted prompt will be influenced by several conditions, such as:
\begin{itemize}
    \item Label Ambiguity: In cases of high label ambiguity, the handcrafted prompt tends to provide more reliable predictions, as soft prompts are difficult to learn effectively in such scenarios.
    \item Dataset Complexity: For datasets with poor CLIP zero-shot performance, reducing reliance on the handcrafted prompt helps avoid steering the learning process in the wrong direction.
    \item Training Epoch: If the soft prompt outperforms the handcrafted prompt as training proceeds, it is better to make more use of the soft prompt, ensuring the most effective predictions are leveraged.
\end{itemize}
Therefore, to further enhance the robustness of our framework against these conditions, a dynamic mixing strategy is adopted to adjust the balancing factor: 
\begin{equation}
    \tilde{p}_{i} = \alpha(t) p(y=i\mid \boldsymbol{x};\boldsymbol{s}) + (1-\alpha(t))p(y=i\mid \boldsymbol{x};\boldsymbol{h}),
    \label{mix_dynamic}
\end{equation}
\begin{equation}
\alpha (t) = \mathrm{min}\{\frac{t}{T'}\lambda, \lambda \},
\end{equation}
where $T'$ controls how quickly the balancing factor grows over time, adjusting the shift between the two outputs' contributions and $\lambda$ sets the maximum weight that the output of soft prompts can have, ensuring its influence is capped.
Together, they dynamically balance the contributions of the two prompts for robustness.

\section{Experiments}
\label{sec:experiment}
\begin{table*}[!htbp]
\caption{
Performance comparison of vanilla prompt learning and our framework. We evaluate the test accuracy of these two methods for different PLL training objectives at different label ambiguities. Supervised means prompt learning with true labels. 
The standard deviation is shown in parentheses. 
}
\label{tab: main}
\centering
\resizebox{1.00\textwidth}{!}{
\setlength{\tabcolsep}{1mm}{
\begin{tabular}{c|cccccccccccc} 
\toprule
\multicolumn{1}{l|}{}       & \multicolumn{3}{c|}{Caltech101}                  & \multicolumn{3}{c|}{DTD}                         & \multicolumn{3}{c|}{FGVCAircraft}                & \multicolumn{3}{c}{ImageNet}                    \\ 
\hline
\multirow{2}{*}{Supervised} & \multicolumn{3}{c|}{92.03}                       & \multicolumn{3}{c|}{62.95}                       & \multicolumn{3}{c|}{27.27}                       & \multicolumn{3}{c}{60.90}                         \\
                            & \multicolumn{3}{c|}{(0.22)}                        & \multicolumn{3}{c|}{(0.43)}                        & \multicolumn{3}{c|}{(3.01)}                        & \multicolumn{3}{c}{(0.64)}                          \\ 
\hline
Zero-shot                   & \multicolumn{3}{c|}{85.84}                       & \multicolumn{3}{c|}{42.79}                       & \multicolumn{3}{c|}{17.07}                       & \multicolumn{3}{c}{58.16}                         \\ 
\hline
$q$                         & 0.1            & 0.3            & 0.5            & 0.1            & 0.3            & 0.5            & 0.1            & 0.3            & 0.5            & 0.1            & 0.3            & 0.5             \\ 
\hline
\multirow{2}{*}{PRODEN}     & 91.20          & 88.78          & 82.05          & 60.85          & 54.70          & 42.15          & 21.03          & 9.15           & 5.58           & 60.20& 47.03& 10.18\\
                            & (0.37)         & (0.11)         & (1.37)          & (0.42)          & (1.35)          & (1.23)          & (1.65)          & (2.11)          & (0.97)          & (0.19)& (2.26)& (0.19)\\
\multirow{2}{*}{+ours}      & \textbf{91.98} & \textbf{91.38} & \textbf{90.55} & \textbf{61.95} & \textbf{56.40} & \textbf{52.77} & \textbf{24.83} & \textbf{21.27} & \textbf{19.03} & \textbf{61.90 }& \textbf{61.55}& \textbf{60.77}\\
                            & (0.26)          & (0.39)          & (0.32)          & (1.20)          & (0.49)          & (2.01)          & (1.06)          & (0.93)          & (0.37)          & (0.27)& (0.30)& (0.11)\\ 
\hline
\multirow{2}{*}{CC}         & 91.68          & 91.23          & \textbf{90.83} & 61.10          & 55.58          & 48.95          & 26.00          & \textbf{22.42} & 18.65          & 61.40& 61.08& 60.83\\
                            & (0.57)          & (0.38)          & (0.66)          & (0.33)          & (1.25)          & (0.99)          & (0.75)          & (0.66)          & (0.15)          & (0.53)& (0.72)& (0.69)\\
\multirow{2}{*}{+ours}      & \textbf{92.35} & \textbf{91.52} & 90.80          & \textbf{61.25} & \textbf{56.25} & \textbf{51.77} & \textbf{26.00} & 22.00          & \textbf{19.85} & \textbf{62.12}& \textbf{61.50 }& \textbf{61.08}\\
                            & (0.27)          & (0.19)          & (0.43)          & (0.56)          & (0.60)          & (0.94)          & (0.89)          & (0.56)          & (0.34)          & (0.11)& (0.32)& (0.24)\\ 
\hline
\multirow{2}{*}{LW}         & 91.35          & 89.02          & 82.97          & 61.65          & 54.73          & 41.77          & 20.77          & 9.22           & 5.85           & 60.30& 45.50& 8.80\\
                            & (0.26)          & (0.37)          & (1.13)          & (1.11)          & (1.84)          & (1.36)          & (1.15)          & (2.04)          & (0.93)          & (0.82)& (3.52)& (1.58)\\
\multirow{2}{*}{+ours}      & \textbf{92.08} & \textbf{91.38} & \textbf{90.67} & \textbf{61.70} & \textbf{56.38} & \textbf{52.95} & \textbf{25.50} & \textbf{21.65} & \textbf{19.43} & \textbf{62.10}& \textbf{61.65}& \textbf{60.80}\\
                            & (0.37)          & (0.29)          & (0.16)          & (0.47)          & (0.94)          & (1.65)          & (0.75)          & (0.84)          & (0.22)          & (0.35)& (0.46)& (0.14)\\ 
\hline
\multirow{2}{*}{PLLCR}      & 91.67          & 91.68          & \textbf{91.10} & \textbf{62.38} & 58.25          & 49.60          &       24.62         &    14.60            & 8.80           & 60.32& 59.60& 59.20\\
                            & (0.29)          & (0.31)          & (0.25)          & (0.38)          & (0.93)          & (1.64)          &       (0.81)         &      (1.99)          & (2.43)          & (0.31)& (0.58)& (3.32)\\
\multirow{2}{*}{+ours}      & \textbf{91.95} & \textbf{91.75} & 91.05          & 62.03          & \textbf{58.50} & \textbf{52.05} & \textbf{26.27}          & \textbf{21.68 }         & \textbf{19.10} & \textbf{61.87}& \textbf{60.77}& \textbf{60.03}\\
                            & (0.25)          & (0.34)          & (0.59)          & (1.09)          & (0.94)          & (0.84)          & (0.48)          & (0.20)          & (0.30)          & (0.24)& (0.21)& (0.42)\\ 
\hline
\multirow{2}{*}{PiCO}       & 90.47          & 88.75          & 87.75          & 61.70          & 55.08          & 48.08          & 23.60          & 18.27          & 14.75          & 58.98& 55.35& 52.72\\
                            & (0.19)          & (0.74)          & (0.76)          & (1.07)          & (1.22)          & (1.58)          & (2.16)          & (2.83)          & (3.11)          & (0.59)& (0.35)& (0.93)\\
\multirow{2}{*}{+ours}      & \textbf{91.92} & \textbf{91.33} & \textbf{90.72} & \textbf{62.67} & \textbf{58.42} & \textbf{55.08} & \textbf{25.12} & \textbf{22.18} & \textbf{20.32} & \textbf{62.03}& \textbf{60.92}& \textbf{59.80}\\
                            & (0.11)          & (0.19)          & (0.15)          & (0.92)          & (0.41)          & (0.93)          & (1.07)          & (1.00)          & (0.71)          & (0.25)& (0.19)& (0.37)\\ 
\hline
\multirow{2}{*}{CAVL}       & 91.67          & 90.97          & 88.62          & \textbf{60.85} & 50.38          & 39.80          & 24.20          & 7.05           & 1.88           & 60.33& 58.88& 57.55\\
                            & (0.40)          & (0.51)          & (1.30)          & (0.97)          & (1.45)          & (4.97)          & (0.51)          & (3.50)          & (0.79)          & 0.62& 0.94& (1.18)\\
\multirow{2}{*}{+ours}      & \textbf{91.82} & \textbf{91.58} & \textbf{90.73} & 60.80          & \textbf{57.18} & \textbf{51.82} & \textbf{25.25} & \textbf{21.45} & \textbf{18.85} & \textbf{62.00}& \textbf{61.15}& \textbf{60.22}\\
                            & (0.33)          & (0.43)          & (0.13)          & (0.95)          & (0.45)          & (1.67)          & (0.97)          & (0.47)          & (0.17)          & (0.19)& (0.25)& (0.39)\\
\bottomrule
\end{tabular}
}
}
\resizebox{1.00\textwidth}{!}{
\setlength{\tabcolsep}{1mm}{
\begin{tabular}{c|cccccccccccc} 
\toprule
\multicolumn{1}{l|}{}       & \multicolumn{3}{c|}{OxfordPets}                  & \multicolumn{3}{c|}{StanfordCars}                & \multicolumn{3}{c|}{UCF101}                      & \multicolumn{3}{c}{Food101}                       \\ 
\hline
\multirow{2}{*}{Supervised} & \multicolumn{3}{c|}{87.82}                       & \multicolumn{3}{c|}{71.52}                       & \multicolumn{3}{c|}{76.55}                       & \multicolumn{3}{c}{77.15}                         \\
                            & \multicolumn{3}{c|}{(0.43)}                        & \multicolumn{3}{c|}{(0.78)}                        & \multicolumn{3}{c|}{(0.42)}                        & \multicolumn{3}{c}{(0.17)}                          \\ 
\hline
Zero-shot                   & \multicolumn{3}{c|}{85.69}                       & \multicolumn{3}{c|}{55.82}                       & \multicolumn{3}{c|}{61.88}                       & \multicolumn{3}{c}{77.39}                         \\ 
\hline
$q$                         & 0.1            & 0.3            & 0.5            & 0.1            & 0.3            & 0.5            & 0.1            & 0.3            & 0.5            & 0.1            & 0.3            & 0.5             \\ 
\hline
\multirow{2}{*}{PRODEN}     & 87.20          & 86.30          & 79.08          & 60.65          & 12.10          & 9.53           & 73.38          & 63.23          & 44.83          & 76.75          & 76.17          & 73.58           \\
                            & (0.45)         & (0.70)         & (1.52)         & (1.03)         & (1.88)         & (0.66)         & (0.77)         & (0.86)         & (0.76)         & (0.54)         & (0.49)         & (0.47)          \\
\multirow{2}{*}{+ours}      & \textbf{88.17} & \textbf{88.25} & \textbf{88.15} & \textbf{68.20} & \textbf{62.25} & \textbf{57.70} & \textbf{74.58} & \textbf{70.72} & \textbf{68.12} & \textbf{77.85} & \textbf{78.40} & \textbf{78.52}  \\
                            & (0.37)         & (0.92)         & (0.74)         & (0.90)         & (1.03)         & (1.06)         & (0.73)         & (0.59)         & (0.98)         & (0.23)         & (0.23)         & (0.31)          \\ 
\hline
\multirow{2}{*}{CC}         & 87.97          & 87.88          & 87.45          & \textbf{70.10} & \textbf{66.73} & \textbf{62.73} & 74.67          & 71.47          & 68.95          & 76.75          & 76.17          & 75.45           \\
                            & (0.62)         & (0.31)         & (0.27)         & (0.60)         & (0.26)         & (1.16)         & (0.26)         & (0.35)         & (1.46)         & (0.54)         & (0.49)         & (0.92)          \\
\multirow{2}{*}{+ours}      & \textbf{88.55} & \textbf{88.65} & \textbf{88.62} & 70.03          & 65.85          & 62.42          & \textbf{75.15} & \textbf{71.60} & \textbf{69.73} & \textbf{77.60} & \textbf{77.80} & \textbf{78.05}  \\
                            & (0.45)         & (0.68)         & (0.33)         & (0.45)         & (0.95)         & (0.91)         & (0.56)         & (0.48)         & (0.63)         & (0.36)         & (0.23)         & (0.25)          \\ 
\hline
\multirow{2}{*}{LW}         & 87.42          & 86.18          & 79.85          & 60.27          & 12.52          & 10.85          & 73.05          & 62.28          & 44.75          & 77.47          & 76.92          & 73.05           \\
                            & (0.41)          & (0.91)          & (1.20)          & (0.28)          & (2.11)          & (2.16)          & (1.28)          & (0.68)          & (0.69)          & (0.44)          & (0.61)          & (0.73)           \\
\multirow{2}{*}{+ours}      & \textbf{88.07} & \textbf{88.50} & \textbf{87.90} & \textbf{68.75} & \textbf{62.25} & \textbf{57.95} & \textbf{74.67} & \textbf{70.70} & \textbf{68.10} & \textbf{77.88} & \textbf{78.10} & \textbf{78.45}  \\
                            & (0.46)          & (0.80)          & (0.75)          & (0.81)          & (0.75)          & (0.98)          & (0.59)          & (0.60)          & (0.74)          & (0.26)          & (0.34)          & (0.32)           \\ 
\hline
\multirow{2}{*}{PLLCR}      & \textbf{86.70} & 86.78          & 85.62          & 66.25          & 61.33          & 25.05                & \textbf{74.68} & 70.22          & 65.20          & 76.03          & 75.70          & 73.95           \\ 
                            & (0.76)          & (0.35)          & (0.94)          & (0.70)          & (0.33)          &    (0.70)            & (0.58)          & (0.65)          & (1.00)          & (0.50)          & (0.42)          & (0.67)           \\
\multirow{2}{*}{+ours}      & 86.52          & \textbf{87.27} & \textbf{87.27} & \textbf{67.05} & \textbf{64.55} & \textbf{59.85} & 74.20          & \textbf{72.22} & \textbf{69.95} & \textbf{76.12} & \textbf{76.22} & \textbf{76.53}  \\
                            & (0.57)          & (0.66)          & (0.47)          & (0.38)          & (0.73)          & (1.57)          & (0.31)          & (0.69)          & (1.04)          & (0.45)          & (0.47)          & (0.48)           \\ 
\hline
\multirow{2}{*}{PiCO}       & 85.90          & 81.70          & 78.40          & 64.27          & 55.15          & 49.95          & 73.45          & 67.23          & 64.35          & 74.03          & 70.22          & 67.97           \\
                            & (1.49)          & (1.07)          & (1.86)          & (0.73)          & (0.97)          & (1.51)          & (0.55)          & (0.69)          & (0.73)          & (0.40)          & (1.01)          & (1.08)           \\
\multirow{2}{*}{+ours}      & \textbf{87.65} & \textbf{86.78} & \textbf{87.15} & \textbf{68.55} & \textbf{63.67} & \textbf{59.55} & \textbf{74.80} & \textbf{71.65} & \textbf{69.55} & \textbf{76.65} & \textbf{76.05} & \textbf{75.68}  \\
                            & (1.22)          & (1.34)          & (1.42)          & (0.59)          & (0.81)          & (0.67)          & (0.98)          & (0.77)          & (0.38)          & (0.39)          & (0.61)          & (0.53)           \\ 
\hline
\multirow{2}{*}{CAVL}       & 87.83          & 87.88          & 86.03          & 69.65          & 63.75          & 57.35          & 73.75          & 69.30          & 63.77          & 76.90          & 76.75          & 75.53           \\
                            & (0.70)          & (1.13)          & (0.85)          & (0.26)          & (0.43)          & (1.27)          & (1.02)          & (1.12)          & ((2.89)         & (0.32)          & (0.17)          & (0.43)           \\
\multirow{2}{*}{+ours}      & \textbf{88.82} & \textbf{88.85} & \textbf{88.88} & \textbf{69.85} & \textbf{65.47} & \textbf{61.65} & \textbf{74.57} & \textbf{71.78} & \textbf{69.02} & \textbf{77.50} & \textbf{78.00} & \textbf{78.10}  \\
                            & (0.48)          & (0.27)          & (0.25)          & (0.61)          & (1.15)          & (0.78)          & (0.63)          & (0.48)          & (0.78)          & (0.16)          & (0.16)          & (0.23)           \\
\bottomrule
\end{tabular}
}
}
\end{table*}

\subsection{Experimental Setting}
\textbf{Datasets.}\quad
\label{dataset}
We adopt $8$ image recognition datasets: ImageNet \cite{deng2009imagenet} and Caltech101 \cite{fei2004learning} for generic object classification; OxfordPets \cite{parkhi2012cats}, StanfordCars \cite{krause20133d}, FGVCAircraft \cite{maji2013fine}, and Food101 \cite{bossard2014food} for fine-grained classification; UCF101 \cite{soomro2012ucf101} for action classification; DTD \cite{cimpoi2014describing} for texture classification.
In order to evaluate the model's performance for partial-label learning, as the same in Section \ref{sec:preliminaries:pll}, we define the level of label ambiguity $q$ as the uniform probability of flipping negative labels $\Bar{y}_i \ne y_i$ to false-positive labels inside the candidate label set $Y_i$: $q = \mathrm{Pr}(\bar{y}_i \in Y_i |\bar{y}_i \ne y_i )$. 
In addition, we use a 16-shot fine-tuning strategy, randomly choosing $16$ images per class from the partialized dataset.

\noindent\textbf{Implementation Details.}\quad 
Our implementation is based on Pytorch \cite{paszke2019pytorch}.
We apply prompt learning on a pre-trained CLIP whose backbone of the image encoder is ResNet-50 \cite{he2016resnet}.
The total number of the learnable prompt tokens is $16$, and the fixed class tokens are at the end of the prompt.
Models are trained with a batch size of $32$ and $50$ total epochs for each method and dataset, except for ImageNet, which sets the batch size to $256$. The optimizer is SGD with a cosine decay schedule annealing the learning rate to $0.00001$. The learning rate for prompt learning and our framework is initialized to be $0.002$. 
Following CoOp\cite{zhou2022coop}, the learnable vectors are initialized from a zero-mean Gaussian distribution with a standard deviation equal to 0.02. As for the handcrafted prompts of our framework, we follow the result of prompt engineering of CLIP \cite{radford2021clip}. 
For example, for the dataset of generic object classification, we adopt ``a photo of a \textless{}CLS\textgreater{}.'' as the handcrafted prompt. For the fine-grained dataset, we adopt the prompts that specify the category, such as ``a photo of a \textless{}CLS\textgreater{}, a type of food.'' for Food101. 
For the hyperparameters of our method, we set $\lambda = 0.5$, $T' = 25$, $\beta = 1$. Moreover, if a confidence matrix is required in the PLL method, it will be initialized with the model output before training. We conduct all experiments on eight NVIDIA RTX 3090 GPUs and report the average test accuracy and the standard deviation of 4 experiments with the seeds fixed. 

\noindent\textbf{Training Objectives.}\quad
We prove the effectiveness of our framework by incorporating six state-of-the-art PLL methods: 
PRODEN \cite{lv2020progressive}, CC \cite{feng2020provably}, LW \cite{wen2021leveraged}, PiCO \cite{wang2021pico}, PLLCR \cite{wu2022revisiting} and CAVL \cite{zhang2021exploiting}. 
\begin{itemize}
    \item PRODEN \cite{lv2020progressive} progressively identifies the true label and seamlessly updates the model parameters.
    \item  CC \cite{feng2020provably} assumes that the generation process of candidate labels is uniform and is provably classifier-consistent.
    \item  LW \cite{wen2021leveraged} weights the loss function by considering the trade-off between losses on candidate and non-candidate labels. We use the loss function version of LWC (LW-Cross entropy) and set the weighting parameter: $\alpha=\beta=1$.
    \item PiCO \cite{wang2021pico} divides the learning process into two components: representation learning by contrastive loss and label disambiguation by prototype and pseudo target updating. We set the queue size that stores key embeddings to be $1024$ and the prototype updating parameter $\gamma = 0.9$. In addition, we disable contrastive learning only for one epoch since the classifier already has a satisfying initialization at the start.
    \item PLLCR \cite{wu2022revisiting} revisits the idea of consistency regularization and performs supervised learning on non-candidate labels and consistency regularization on candidate labels. We set $T^{'} = 50$ in the dynamic balancing strategy: $\gamma(t)=\min \left\{\frac{t}{T^{\prime}} \lambda, \lambda\right\}$. Moreover, We set $T^{'}$ half of the total epochs, which is the same as the original setting in PLLCR \cite{wu2022revisiting}.
    \item CAVL \cite{zhang2021exploiting} utilizes class activation value (CAV) learning that selects the true label with the maximum CAV. 
\end{itemize}

\subsection{Main Results}

In Table \ref{tab: main}, we compare our framework with vanilla prompt learning for six PLL training objectives on ten benchmark datasets when $q= \{0.1, 0.3, 0.5\}$. 
For supervised learning, we use the same settings as vanilla prompt learning, except the cross entropy is used, while for CLIP zero-shot, we use the same prompt as the handcrafted prompt in our framework.
It is shown that the performance of vanilla prompt learning drops significantly with higher label ambiguity and sometimes even under-performs zero-shot inference, verifying its lack of robustness.
On the contrary, our framework not only shows robustness with highly ambiguous candidate labels but also improves general performance that matches supervised learning.
For instance, in OxfordPets with LW at $q = 0.5$, vanilla prompt learning under-performs zero-shot inference by 5.84\%, while our framework outperforms supervised learning by 2.21\%.
Moreover, our framework demonstrates consistent improvement across all levels of label ambiguity, not just in cases of high ambiguity. For example, in the Caltech101 dataset, our framework outperforms vanilla prompt learning in 16 out of 18 cases, regardless of the level of label ambiguity and PLL training objectives. Particularly, the more label ambiguity, the more gain in performance with our framework.

In addition, in some datasets, the test accuracy of our framework will increase with more label ambiguity, like in Food101 and OxfordPets. We believe these two fine-grained datasets contain considerable label noise \cite{bossard2014food,parkhi2012cats}, which is supported by the result in CoOp \cite{zhou2022coop}, so more label ambiguity actually makes the true label more likely in the candidate label set. Since our framework can better deal with highly ambiguous candidate labels, more ambiguous candidate labels have exerted a relatively more positive effect on our framework.

There are a few cases where prompt learning with CC and PLLCR has outperformed our method. We conjecture that because these two PLL objectives are considered average-based, whose prediction is not based on the prediction from the last epoch, they suffer less from the error accumulation problem and will not benefit from our framework as effectively as other training objectives.

The main result has verified the efficacy of our framework. 
By aligning with the handcrafted prompt, the model can make more accurate predictions at the initial stage of the learning process. Moreover, the class posterior predicted by the handcrafted prompt is hardly affected by the level of label ambiguity, making the learning process more robust.

\vspace{-0.3cm}
\subsection{Ablation Studies}
\vspace{-0.3cm}

In this part, we conduct experiments to assess the effectiveness of our framework with different handcrafted prompts, learnable context lengths, visual backbones, and balancing strategies in 3 datasets: DTD, FGVCAircraft and Caltech101. 

\noindent\textbf{Impact of Different Handcrafted Prompts.} \quad 
Because our framework incorporates handcrafted prompts, it is crucial to determine the performance of our framework with differently crafted prompts. In Table \ref{tab:sensitivity_hc}, we design some prompts with zero-shot performance lower than the default handcrafted prompts \cite{radford2021clip} and evaluate our framework with these prompts at different levels of label ambiguity. 
\textit{Introducing a handcrafted prompt can consistently improve the performance, regardless of the quality of the prompt.} Better handcrafted prompts have resulted in better test accuracy in our experiments, which indicates that the handcrafted prompts indeed guide the fine-tuning process with candidate labels. Nonetheless, the performance is improved by a large margin even when we simply set the handcrafted prompt ``\textless{}CLS\textgreater{}.''. 

\begin{table*}[!htbp]
\caption{Performance comparison of our framework for different handcrafted prompts.}

\label{tab:sensitivity_hc}
\tiny
\centering
\resizebox{1.00\textwidth}{!}{
\setlength{\tabcolsep}{2mm}{
\begin{tabular}{c|c|c|c|c|c} 
\toprule
\multirow{2}{*}{Dataset}                                                & \multirow{2}{*}{Handcrafted Prompt}                                                         & \multicolumn{3}{c|}{$q$}                                                    & \multicolumn{1}{c}{\multirow{2}{*}{\begin{tabular}[c]{@{}c@{}}CLIP- \\Zeroshot\end{tabular}}}  \\
\cmidrule{3-5}
                                                                        &                                                                                             & 0.1                     & 0.3                     & 0.5                     & \multicolumn{1}{c}{}                                                                           \\ 
\midrule
\multirow{4}{*}{DTD}                                                    & \begin{tabular}[c]{@{}c@{}}without handcrafted prompt\\~(only prompt learning)\end{tabular} & 61.70$\pm$1.07          & 55.08$\pm$1.22          & 48.08$\pm$1.58          & -                                                                                               \\ 
\cmidrule{2-2}
                                                                        & “a photo of a \textless{}CLS\textgreater{}.”                                                                         & 62.15$\pm$0.63          & 56.85$\pm$0.63          & 53.20$\pm$0.91          & 40.36                                                                                           \\ 
\cmidrule{2-2}
                                                                        & “\textless{}CLS\textgreater{}.”                                                                                      & 62.45$\pm$0.18          & 56.67$\pm$0.98          & 53.23$\pm$1.20          & 41.13                                                                                           \\ 
\cmidrule{2-2}
                                                                        & “\textless{}CLS\textgreater{} texture.”                                                                              & \textbf{62.67$\pm$0.92} & \textbf{58.42$\pm$0.41} & \textbf{55.08$\pm$0.93} & \textbf{42.79}                                                                                  \\ 
\midrule
\multirow{4}{*}{\begin{tabular}[c]{@{}c@{}}FGVC\\Aircraft\end{tabular}} & \begin{tabular}[c]{@{}c@{}}without handcrafted prompt\\~(only prompt learning)\end{tabular} & 23.60$\pm$2.16          & 18.27$\pm$2.83          & 14.75$\pm$3.11          & -                                                                                               \\ 
\cmidrule{2-2}
                                                                        & “a photo of a \textless{}CLS\textgreater{}.”                                                                         & 24.70$\pm$1.16          & 21.82$\pm$1.42          & 19.60$\pm$0.81          & 15.84                                                                                           \\ 
\cmidrule{2-2}
                                                                        & “\textless{}CLS\textgreater{}.”                                                                                      & 24.65$\pm$0.89          & 20.85$\pm$0.84          & 19.02$\pm$0.71          & 15.54                                                                                           \\ 
\cmidrule{2-2}
                                                                        & \begin{tabular}[c]{@{}c@{}}“a photo of a \textless{}CLS\textgreater{}, \\a type of aircraft.”\end{tabular}           & \textbf{25.12$\pm$1.07} & \textbf{22.18$\pm$1.00} & \textbf{20.32$\pm$0.71} & \textbf{17.07}                                                                                  \\ 
\midrule
\multirow{3}{*}{Caltech101}                                             & \begin{tabular}[c]{@{}c@{}}without handcrafted prompt \\(only prompt learning)\end{tabular} & 90.47$\pm$0.19          & 88.75$\pm$0.74          & 87.75$\pm$0.76          & -                                                                                               \\ 
\cmidrule{2-2}
                                                                        & “\textless{}CLS\textgreater{}.”                                                                                      & 91.42$\pm$0.19          & 90.97$\pm$0.51          & 90.05$\pm$0.73          & 81.34                                                                                           \\ 
\cmidrule{2-2}
                                                                        & “a photo of a \textless{}CLS\textgreater{}.”                                                                         & \textbf{91.92$\pm$0.11} & \textbf{91.33$\pm$0.19} & \textbf{90.72$\pm$0.15} & \textbf{85.84}                                                                                  \\
\bottomrule
\end{tabular}
}
}
\vspace{-0.6cm}
\end{table*}

\noindent\textbf{Impact of Different Learnable Context Lengths.} \quad We evaluate our framework when $q = 0.3$ while varying the learnable context length of the learnable prompt from $1$ to $16$. The result is shown in Table \ref{tab: token_len}. \textit{Our framework is superior with different learnable context lengths.} It enhances the performance by a large margin with any of the selected context lengths. \textit{The longer, the better.} Both methods perform worst when the learnable context length is 1 and improve with a longer length. We conjecture that it is because the neural network with fewer parameters is better calibrated \cite{guo2017calibration}, reducing the contribution of the model's zero-shot ability to help disambiguate the candidate labels.

\begin{table*}[!h]
\centering
\vspace{-0.6cm}
\caption{Performance comparison with different context lengths when $q=0.3$.}
\label{tab: token_len}
\resizebox{1.00\textwidth}{!}{
\setlength{\tabcolsep}{5mm}{
\begin{tabular}{c|cccccc} 
\toprule
\multirow{2}{*}{\begin{tabular}[c]{@{}c@{}}Context\\Length\end{tabular}} & \multicolumn{2}{c|}{DTD}                                                & \multicolumn{2}{c|}{FGVCAircraft}                                       & \multicolumn{2}{c}{Caltech101}                                           \\ 
\cmidrule{2-7}
                                                                         & PiCO                        & \multicolumn{1}{c|}{PiCO+ours}            & PiCO                        & \multicolumn{1}{c|}{PiCO+ours}            & PiCO                        & PiCO+ours                                  \\ 
\midrule
1                                                                        & 52.20$\pm$1.53 & \textbf{\textbf{54.73$\pm$1.21}}          & 19.70$\pm$0.85 & \textbf{\textbf{21.08$\pm$0.59}}          & 88.92$\pm$0.40 & \textbf{\textbf{90.75$\pm$0.32}}           \\
2                                                                        & 54.50$\pm$0.31 & \textbf{\textbf{57.17$\pm$0.97}}          & 19.88$\pm$0.75 & \textbf{\textbf{21.12$\pm$0.33}}          & 89.33$\pm$0.29 & \textbf{\textbf{90.70$\pm$0.16}}           \\
4                                                                        & 52.12$\pm$2.29 & \textbf{\textbf{57.62$\pm$0.84}}          & 19.52$\pm$0.80 & \textbf{\textbf{21.60$\pm$0.80}}          & 88.85$\pm$0.26 & \textbf{\textbf{90.75$\pm$0.30}}           \\
8                                                                        & 55.30$\pm$0.37 & \textit{\textbf{\textbf{58.45$\pm$0.84}}} & 20.40$\pm$1.11 & \textbf{\textbf{22.05$\pm$0.17}}          & 89.07$\pm$0.31 & \textbf{\textbf{91.03$\pm$0.39}}           \\
16                                                                       & 55.08$\pm$1.22 & \textbf{\textbf{58.42$\pm$0.41}}          & 18.27$\pm$2.83 & \textit{\textbf{\textbf{22.18$\pm$1.00}}} & 88.75$\pm$0.74 & \textit{\textbf{\textbf{91.33$\pm$0.19}}}  \\
\bottomrule
\end{tabular}
}
}
\vspace{-0.6cm}
\end{table*}

\noindent\textbf{Impact of Different Visual Backbones.} \quad We test the effectiveness of our framework with both ResNet-101 and ViT-B/32 across different label ambiguities. The result is shown in Table \ref{tab: arch}. 
\textit{Our framework outperforms vanilla prompt learning with different visual encoders.} Both methods perform better with a better backbone but our framework is more desirable across different backbones. Moreover, the conclusions from the pilot experiment also apply to VLMs with Vision Transformer as the image encoder.

\begin{table*}[!h]
\vspace{-0.6cm}
\caption{Performance comparison with different visual backbones.}
\label{tab: arch}
\resizebox{1.00\textwidth}{!}{
\setlength{\tabcolsep}{0.3mm}{
\begin{tabular}{c|ccccccccc} 
\toprule
              & \multicolumn{3}{c|}{DTD}                                                                                                                                     & \multicolumn{3}{c|}{FGVCAircraft}                                                                                                                            & \multicolumn{3}{c}{Caltech101}                                                                                                                                \\ 
\midrule
$q$           & 0.1                                                & 0.3                                                & \multicolumn{1}{c|}{0.5}                           & 0.1                                                & 0.3                                                & \multicolumn{1}{c|}{0.5}                           & 0.1                                                & 0.3                                                & 0.5                                                 \\ 
\midrule
\multicolumn{10}{c}{\textit{ResNet-101}}                                                                                                                                                                                                                                                                                                                                                                                                                                    \\ \midrule
CLIP-Zeroshot & \multicolumn{3}{c}{37.82}                                                                                                                                    & \multicolumn{3}{c}{18.12}                                                                                                                                    & \multicolumn{3}{c}{90.06}                                                                                                                                     \\
PiCO          & 63.70$\pm$1.34                                     & 56.80$\pm$1.50                                     & 46.40$\pm$1.54                                     & 27.95$\pm$0.42                                     & 17.38$\pm$3.08                                     & 18.00$\pm$4.91                                     & 93.45$\pm$0.42                                     & 92.22$\pm$0.62                                     & 91.03$\pm$0.97                                      \\
PiCO+ours     & \textbf{\textbf{\textbf{\textbf{64.55$\pm$0.75}}}} & \textbf{\textbf{\textbf{\textbf{59.40$\pm$0.70}}}} & \textbf{\textbf{\textbf{\textbf{54.45$\pm$1.92}}}} & \textbf{\textbf{\textbf{\textbf{28.83$\pm$0.64}}}} & \textbf{\textbf{\textbf{\textbf{25.20$\pm$0.19}}}} & \textbf{\textbf{\textbf{\textbf{22.10$\pm$0.37}}}} & \textbf{\textbf{\textbf{\textbf{94.18$\pm$0.29}}}} & \textbf{\textbf{\textbf{\textbf{93.40$\pm$0.44}}}} & \textbf{\textbf{\textbf{\textbf{93.12$\pm$0.38}}}}  \\ 
\midrule
\multicolumn{10}{c}{\textit{ViT-B/32}}                                                                                                                                                                                                                                                                                                                                                                                                                                          \\ \midrule
CLIP-Zeroshot & \multicolumn{3}{c}{44.14}                                                                                                                                    & \multicolumn{3}{c}{19.26}                                                                                                                                    & \multicolumn{3}{c}{91.40}                                                                                                                                     \\
PiCO          & 63.02$\pm$1.04                                     & 56.40$\pm$0.70                                     & 46.75$\pm$1.83                                     & 27.30$\pm$0.80                                     & 23.23$\pm$0.90                                     & 18.82$\pm$1.66                                     & 94.55$\pm$0.42                                     & 93.47$\pm$0.42                                     & 92.95$\pm$0.44                                      \\
PiCO+ours     & \textbf{\textbf{\textbf{\textbf{64.50$\pm$0.27}}}} & \textbf{\textbf{\textbf{\textbf{60.18$\pm$1.10}}}} & \textbf{\textbf{\textbf{\textbf{56.77$\pm$1.01}}}} & \textbf{\textbf{\textbf{\textbf{28.18$\pm$0.19}}}} & \textbf{\textbf{\textbf{\textbf{24.30$\pm$0.29}}}} & \textbf{\textbf{\textbf{\textbf{22.30$\pm$0.48}}}} & \textbf{\textbf{\textbf{\textbf{95.22$\pm$0.15}}}} & \textbf{\textbf{\textbf{\textbf{94.70$\pm$0.19}}}} & \textbf{\textbf{\textbf{\textbf{94.17$\pm$0.38}}}}  \\
\bottomrule
\end{tabular}
}
}
\vspace{-0.6cm}
\end{table*}

\noindent\textbf{Effectiveness of the Dynamic Mixing Strategy.} \quad This experiment is conducted to evaluate the efficacy of our dynamic and mixing strategy. $\alpha(t)$ is set using the same hyperparameters as the main experiment. The result is shown in Table \ref{tab: alpha}. \textit{The dynamic mixing strategy optimizes the model's performance.} 
The model with dynamic technique outperforms all others in 6 and ranked second in 2, out of 9 cases, demonstrating its effectiveness. When only one handcrafted or learnable prompt is leveraged, the model's performance declines and cannot surpass the performance achieved when $\alpha = 0.7$. This validates the rationality of the mixing strategy. 
Lastly, when $\alpha = 0$, it outperforms models with only learnable prompt guidance ($\alpha = 1$) or no guidance (vanilla prompt learning) in most cases, which, to some extent, justifies the effectiveness of explicitly adopting the handcrafted prompt to guide the learning process.

\begin{table*}[!h]
\centering
\vspace{-0.6cm}
\caption{Performance comparison of our framework for different $\alpha$.}
\label{tab: alpha}
\resizebox{1.00\textwidth}{!}{
\setlength{\tabcolsep}{1mm}{
\begin{tabular}{c|ccccccccc} 
\toprule
            & \multicolumn{3}{c|}{DTD}                                                                               & \multicolumn{3}{c|}{FGVCAircraft}                                                                      & \multicolumn{3}{c}{Caltech101}                                                                          \\ 
\midrule
$q$         & 0.1                              & 0.3                              & \multicolumn{1}{c|}{0.5}         & 0.1                              & 0.3                              & \multicolumn{1}{c|}{0.5}         & 0.1                              & 0.3                              & 0.5                               \\ 
\midrule
$\alpha=0$         & 62.05$\pm$1.17      & 57.12$\pm$0.98      & 53.75$\pm$1.41      & 24.68$\pm$0.83      & 22.10$\pm$0.56      & 19.93$\pm$0.31      & 91.70$\pm$0.23      & 91.20$\pm$0.16      & \textbf{\textbf{90.93$\pm$0.22}}  \\
$\alpha=1$       & 61.83$\pm$0.40      & 56.45$\pm$2.29      & 50.98$\pm$3.08      & 25.15$\pm$0.98      & 18.83$\pm$4.24      & 15.35$\pm$3.76      & 91.62$\pm$0.25      & 89.83$\pm$0.68      & 88.47$\pm$0.37       \\
$\alpha=0.7$       & 62.28$\pm$0.33      & \textbf{\textbf{59.38$\pm$1.43}} & 54.57$\pm$1.14      & \textbf{\textbf{25.77$\pm$0.99}} & 22.02$\pm$0.98      & 19.95$\pm$0.38      & 91.85$\pm$0.23      & 91.03$\pm$0.54      & 90.35$\pm$0.63       \\
$\alpha(t)$ & \textbf{\textbf{62.67$\pm$0.92}} & 58.42$\pm$0.41      & \textbf{\textbf{55.08$\pm$0.93}} & 25.12$\pm$1.07      & \textbf{\textbf{22.18$\pm$1.00}} & \textbf{\textbf{20.32$\pm$0.71}} & \textbf{\textbf{91.92$\pm$0.11}} & \textbf{\textbf{91.33$\pm$0.19}} & 90.72$\pm$0.15       \\
\bottomrule
\end{tabular}
}
}
\vspace{-0.6cm}
\end{table*}

\subsection{Further Analysis}

In this part, we demonstrate the effectiveness of our framework with instance-dependent candidate labels or various VLM tuning methods. Moreover, we also highlight its superiority over Unsupervised Prompt Learning \cite{huang2022upl}.

\noindent\textbf{Robustness to Instance-dependent Candidate Labels.} \quad To better simulate the realistic annotation process, we introduce instance-dependent candidate labels and evaluate the performance of our framework under them. Specifically, we use CLIP with the handcrafted prompt of “a photo of a \textless{}CLS\textgreater{}.” to make the zero-shot inference of each example. Then, we select the pseudo labels with top \{10\%, 30\%, 50\%\} confidence as instance-dependent candidate labels. We define the proportion as $q$. The result is shown in Table \ref{tab: instance-dependent}. \textit{Our framework is more robust with instance-dependent candidate labels.} Both vanilla prompt learning and our framework have experienced a significant performance drop since the instance-dependent candidate labels are more challenging for the model than uniform candidate labels. Nevertheless, our framework is more robust and can maintain stable performance with high label ambiguity.

\begin{table*}[!h]
\vspace{-0.6cm}
\caption{Performance comparison with instance-dependent candidate labels.}
\label{tab: instance-dependent}
\resizebox{1.00\textwidth}{!}{
\setlength{\tabcolsep}{0.3mm}{
\begin{tabular}{c|ccccccccc} 
\toprule
          & \multicolumn{3}{c|}{DTD}                                                                                                                   & \multicolumn{3}{c|}{FGVCAircraft}                                                                                                          & \multicolumn{3}{c}{Caltech101}                                                                                                              \\ 
\midrule
$q$       & 0.1                              & 0.3                                                & \multicolumn{1}{c|}{0.5}                           & 0.1                              & 0.3                                                & \multicolumn{1}{c|}{0.5}                           & 0.1                              & 0.3                                                & 0.5                                                 \\ 
\midrule
PiCO      & 50.15$\pm$0.81      & 36.95$\pm$0.62                        & 34.67$\pm$2.11                        & 14.88$\pm$0.70      & 12.47$\pm$0.53                        & 10.07$\pm$2.66                        & 86.78$\pm$0.74      & 84.50$\pm$0.27                        & 84.57$\pm$0.15                         \\
PiCO+ours & \textbf{\textbf{50.17$\pm$1.01}} & \textbf{\textbf{\textbf{\textbf{46.10$\pm$1.15}}}} & \textbf{\textbf{\textbf{\textbf{45.47$\pm$1.45}}}} & \textbf{\textbf{16.70$\pm$0.25}} & \textbf{\textbf{\textbf{\textbf{16.52$\pm$0.30}}}} & \textbf{\textbf{\textbf{\textbf{16.25$\pm$0.64}}}} & \textbf{\textbf{88.97$\pm$0.50}} & \textbf{\textbf{\textbf{\textbf{88.42$\pm$0.41}}}} & \textbf{\textbf{\textbf{\textbf{88.28$\pm$0.44}}}}  \\
\bottomrule
\end{tabular}
}
}
\vspace{-0.6cm}
\end{table*}

\noindent\textbf{Performance comparison with UPL on candidate labels.} \quad 
Unsupervised Prompt Learning \cite{huang2022upl} (UPL) leverages zero-shot predictions to learn from unsupervised data by optimizing hard pseudo-labels with relatively high confidence for each class. 
While it shares similarities with our framework, UPL does not incorporate candidate labels and fails to address the error accumulation issue inherent in PLL. 
As a result, it demonstrates weaker performance compared to our approach as shown in Table \ref{tab:upl}. 
These findings further highlight the effectiveness of our framework, particularly when handling candidate labels.

\begin{table*}
\centering
\caption{Performance comparison on 8 datasets. We set the $q=0.1$ in our dataset and use PRODEN as the surrogate PLL objective. For UPL, we follow the implementation settings in the \cite{huang2022upl} and set the number of learned samples per class to 16.}
\label{tab:upl}
\resizebox{1.00\textwidth}{!}{
\setlength{\tabcolsep}{8mm}{
\begin{tabular}{lccc} 
\toprule
                             & UPL & PRODEN ($q=0.1$) & PRODEN+Ours ($q=0.1$)  \\ 
\midrule
\textit{Avg.} on 8 datasets & 65.89$\pm$ 0.94     & 69.12$\pm$ 0.83                 & \textbf{70.84$\pm$ 0.67}              \\
\bottomrule
\end{tabular}
}}
\vspace{-0.6cm}
\end{table*}

\noindent\textbf{Performance with Different VLM Tuning Methods.} \quad In addition to prompt learning, we integrate our framework with various VLM tuning methods to assess their robustness improvements with candidate labels. 
Specifically, VPT \cite{jia2022visual} adjusts visual prompts in the shallow layers, CLIP-Adapter \cite{gao2023clipadapter} incorporates an adapter module that blends pre-trained features in a residual manner, and Maple \cite{khattak2023maple} simultaneously tunes both visual and textual prompts.
The visual architecture is ViT/B16 and the PLL training objective is PRODEN.
As shown in Table \ref{tab: new_method}, our framework consistently enhances performance across diverse VLM tuning methods, demonstrating its effectiveness and flexibility.

\begin{table*}[h!]
\centering
\vspace{-0.6cm}

\caption{Performance comparison with different VLM tuning methods on Caltech101.}
\label{tab: new_method}
\resizebox{1.00\textwidth}{!}{
\setlength{\tabcolsep}{12mm}{
\begin{tabular}{cccc} 
\hline
\multicolumn{1}{c|}{Method}       & $q=0.1$            & $q=0.3$            & $q=0.5$             \\ 
\hline
\multicolumn{1}{c|}{Adapter}       & 92.55 $\pm$ 0.15 & 92.40 $\pm$ 0.24 & 91.92 $\pm$ 0.43  \\
\multicolumn{1}{c|}{Adapter+ours}  & \textbf{93.13 $\pm$ 0.12} & \textbf{92.87 $\pm$ 0.05} & \textbf{92.62 $\pm$ 0.15}  \\
\hline
\multicolumn{1}{c|}{VPT}           & 93.52 $\pm$ 0.41 & 92.03 $\pm$ 0.36 & 89.02 $\pm$ 0.68  \\
\multicolumn{1}{c|}{VPT+ours}      & \textbf{93.70 $\pm$ 0.25} & \textbf{92.97 $\pm$ 0.44} & \textbf{92.62 $\pm$ 0.33}  \\
\hline
\multicolumn{1}{c|}{MaPLe}         & 95.00 $\pm$ 0.22 & 94.33 $\pm$ 0.08 & 92.77 $\pm$ 0.13  \\
\multicolumn{1}{c|}{MaPLe+ours}    & \textbf{95.15 $\pm$ 0.09} & \textbf{95.00 $\pm$ 0.10} & \textbf{94.55 $\pm$ 0.09}  \\
\hline
\end{tabular}
}
}
\vspace{-0.6cm}
\end{table*}

\vspace{-0.3cm}
\section{Conclusion}
\label{sec:conclusion}

This work, for the first time, investigated the scenario when tuning vision-language models (VLMs) with candidate labels. Throughout a series of experiments, we empirically demonstrated that prompt learning combined with PLL training objectives in a vanilla way can learn from candidate labels.
However, as the ambiguity of candidate labels increases, its performance is degraded. To alleviate this issue, we proposed a framework that enhances the robustness by aligning the dynamically mixed class posterior of the handcrafted and learnable prompt with the model's output to guide the learning process with candidate labels. Comprehensive experimental results and analysis on multiple benchmark datasets demonstrate the effectiveness of our proposed framework.

%
%
%
\bibliographystyle{splncs04}
\bibliography{main}
\end{document}